# The Consequences of the *Framing* of Machine Learning Risk Prediction Models: Evaluation of Sepsis in General Wards

Simon Meyer Lauritsen,[1,2] Bo Thiesson,[1,3] Marianne Johansson Jørgensen,[4] Anders Hammerich Riis,[1] Ulrick Skipper Espelund,[4,5] Jesper Bo Weile,[6,7] Jeppe Lange[2,4]

[1]*Enversion A/S, Fiskerivej 12, 1st floor, 8000 Aarhus C, Denmark;* [2]*Department of Clinical Medicine, Aarhus University, Aarhus N, Denmark;* [3]*Department of Engineering, Aarhus University, Aarhus C, Denmark;* [4]*Department of Research, Horsens Regional Hospital, Horsens, Denmark; Department of Anesthesiology, Horsens Regional Hospital, Denmark;* [6]*Emergency Department, Horsens Regional Hospital, Denmark;* [7]*Research Center for Emergency Medicine, Aarhus University Hospital, Denmark*

*Corresponding author: Simon Meyer Lauritsen (sla@enversion.dk)*

Word count: 4,108

**Summary box**

**What is already known on this topic:**

Machine learning risk prediction models hold great promise in healthcare technology, but many aspects of the emerging machine learning technology are yet unknown.

This holds true for sepsis prediction, where machine learning methods offer seemingly better predictive performance than traditional early warning score methods.

**What this study adds:**

Our study demonstrates that the consequences of model framing mandate attention from clinicians and developers, as the understanding and reporting of framing are pivotal to the successful development and clinical implementation of future AI technology.

Our study shows opposing interpretations of $SpO_2$ in the prediction of sepsis as a direct consequence of differentially framed models.

The importance of proper problem framing is by no means exclusive to sepsis prediction and applies to most clinical risk prediction models.

Given the substantial variation in framing approaches in previous published risk prediction models, it may not be wise to compare the derived results directly.

Model framing, as defined in this study, should be explicitly reported in future studies on machine learning risk prediction models.

# Abstract

**Objectives**

To evaluate the consequences of the framing of machine learning risk prediction models. We evaluate how framing affects model performance and model learning in four different approaches previously applied in published artificial-intelligence (AI) models.

**Design**

Longitudinal cohort study from September 1, 2012 to December 31, 2018.

**Setting and participants**

We analysed structured secondary healthcare data from 221,283 citizens from four Danish municipalities (Odder, Hedensted, Skanderborg, and Horsens) who were 18 years of age or older. A total of 19,976 inpatient admissions were included from Horsens Regional Hospital.

**Main outcome measures**

Model performance, including the area under the receiver operating characteristic curve and area under the precision recall curve. Shapley additive explanations were used to evaluate the associations learned in the four model framing approaches by aligning individual predictions with decisive inputs.

**Results**

The four models had similar population level performance (a mean area under the receiver operating characteristic curve of 0.73–0.82), in contrast to the mean average precision, which varied greatly from 0.007 to 0.385. Correspondingly, the percentage of missing values also varied between framing approaches. The on-clinical-demand framing, which involved samples for each time the clinicians made an early warning score assessment, showed the lowest percentage of missing values among the vital sign parameters, and this model was also able to learn more temporal dependencies than the others. The Shapley additive explanations demonstrated opposing interpretations of $SpO_2$ in the prediction of sepsis as a consequence of differentially framed models.

**Conclusions**

The profound consequences of framing mandate attention from clinicians and AI developers, as the understanding and reporting of framing are pivotal to the successful development and clinical implementation of future AI technology. Model framing must reflect the expected clinical environment. The importance of proper problem framing is by no means exclusive to sepsis prediction and applies to most clinical risk prediction models.

# Introduction

Machine learning risk-prediction models hold great promise in artificial intelligence (AI) healthcare technology. But very few machine learning models in healthcare are tested prospectively and clinically deployed [1–4]. Many aspects of the emerging AI technology are yet unknown to the end-user, the healthcare professional, who needs to comprehend the AI technology to ensure a genuine clinical translation. The consequences of such aspects of AI healthcare technology need to be addressed and readily integrated into future reports on AI models in healthcare, such as it was recently suggested for the effect of censoring on risk predictions [5].

As an area of research, machine learning risk-prediction models for sepsis is very active, with new methodologies appearing regularly [6–22]. Prediction of sepsis provides a perfect example of the consequences that framing can exert on a developed risk prediction model. Machine learning methods have gained ground on traditional early warning score (EWS) methods by offering seemingly better predictive performance [6,9,12,15,18,21,23–25].

Although the purpose of most sepsis risk prediction methods is the same—to predict the onset of sepsis at the earliest possible time—there are many different approaches to framing a machine learning risk-prediction model. In the context of this study, the framing of machine learning risk-prediction models needs to be understood in this way: it is to predict the specific medical event for a patient, at a certain time point or throughout the hospital admission, using fixed or varying time intervals between each prediction interval.

Most sepsis risk prediction models are trained using data from intensive care units (ICU)[21,26], rather than data from general medical and surgical wards. To some extent, this is due to a very popular ICU dataset from the Massachusetts Institute of Technology, called MIMIC-III [27,28]. In ICUs, up to one-third of all patients have sepsis [29]. Also, in ICUs, vast quantities of data are continuously registered, which provides good conditions for sequential machine learning risk-prediction models, yielding predictions every 5 minutes or so. In contrast, the prevalence of sepsis is much lower in medical and surgical wards, and since the clinical setting differs, so should the framing of the risk prediction model.

Here, paraclinical parameters are registered less frequently and at random intervals, such as blood tests one to three times per day and vital parameters one to eight times per day, depending on the status of the patient. A model developed for use in an ICU should not be applied in other clinical environments, as model deployment in these two clinical environments cannot be equated and could possibly lead to high false-positive rates and associated fatigue of the end-user. Real clinical translation can only be reached if this fatigue is avoided, and the purpose should instead be to capture the development of sepsis—and other critical conditions—early and to initiate the necessary interventions to ensure that the patient does not deteriorate.

The purpose of this study is to evaluate the consequences of the framing of machine learning risk-prediction models. We apply four different framing approaches to the same generic dataset using a sepsis risk-prediction model as an example. We evaluate how framing affects model performance and model learning. Additionally, we argue that problem framing should mimic the intended use of the prediction model and appeal for this framing to be rigorously discussed in future machine learning risk-prediction models.

# Methods

**Data source**

The dataset used in this study came from the CROSS-TRACKS cohort, a population-based, open cohort, containing routinely collected data from primary and secondary healthcare partners, combined with data from national registries. A full description of the cohort has been published [30], but, in brief, the CROSS-TRACKS cohort consists of all citizens aged 18+ residing in the catchment area of Horsens Regional Hospital, a teaching hospital in the Central Denmark region, which serves a population of 221,283 residents in four Danish municipalities (Odder, Hedensted, Skanderborg, and Horsens). The cohort entry date was September, 1 2012, and inclusion will last through 2022. Also, a 10-year look-back period and a 5-year follow-up period are included in the cohort. The cohort offers a complete, multi-dimensional model of primary and secondary healthcare data, including data from municipalities. This multi-dimensionality is possible due to a merger of all the data sets via the unique personal identification number given to all Danish citizens [31].

**Study population**

The study population consisted of all patients hospitalised at Horsens Regional Hospital from September, 1 2012 to December, 31 2018 who were 18 years of age or older. Outpatient contacts were not included. We excluded inpatient admissions shorter than 24 hours or longer than 50 days. In the study period, 19,976 inpatient admissions were applicable (for cohort characteristics see Supplementary Table 2). The included admissions were distributed across 13,134 unique residents. The prevalence of sepsis among these admissions was 6.25%.

**Disease definition**

We defined sepsis for individual cases in the study population by the recent Sepsis-3 [32,33], also evaluated in a machine learning algorithm by Moor et al. [19], according to which both suspected infection and organ dysfunction were required to be present [19,32,33].

Suspected infection was defined in the dataset by the coherent occurrence during the inpatient admission of i) samples obtained (regardless of type) for culture purposes and ii) antibiotic administration. When a sample was obtained for culture before the administration of antibiotics, the antibiotic had to be administered within 72 hours. If the antibiotic was administered first, then the sample for culture had to follow within 24 hours.

The degree of organ dysfunction was determined by the acute increase of two or more points in a Sequential Organ Failure Assessment (SOFA) score [32,34]. In implementing the organ dysfunction criterion, we used a 72-hour window from 48 hours before to 24 hours after the index time of suspected infection, as suggested by Singer et al.[33] and Moor et al.[19] The Sepsis-3 implementation is illustrated in Figure 1.

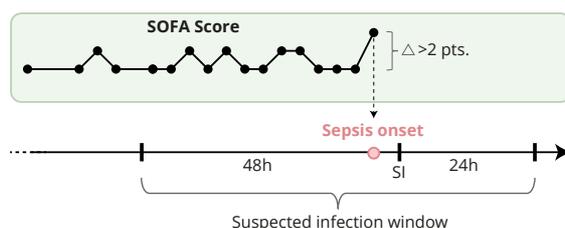

**Figure 1 | Sepsis definition.** Figure showing how sepsis onset is defined. SOFA: Sequential Organ Failure Assessment Scores, SI: Suspected Infection.

**Framing of the prediction model**

Eight different approaches to framing the sepsis risk prediction model are depicted in Figure 2. These approaches were identified in a literature review, as discussed below.

Four approaches, *a–d,* were selected as generic examples that demonstrate the consequences of the framing of the machine learning risk-prediction model. The four chosen framing approaches were selected by the authors because they were believed to represent realistic framings for a clinical deployment in our general ward setup:

- *Fixed time to onset* (Figure 2*a*)*.* The time of prediction for a sepsis positive patient is fixed to some number of hours, such as 12, before sepsis onset. For the sepsis negative patients, the "onset" time is set at a random time during admission.

- *Sliding window* (Figure 2*b*). The entire admission is split into chunks, e.g. 8-hours; each chunk is labelled sepsis positive if sepsis onset occurs within the prediction window and sepsis negative otherwise.

- *Sliding window with dynamic inclusion* (Figure 2*c*). The sampling only starts when some criteria are met, such as when a SOFA score of greater than 0 is registered. This approach only differs from the *sliding window* approach by sampling fewer times, omitting samples known to be negative, and is thereby a way to reduce the number of true negative samples in the dataset without reducing the true positives.

- *On clinical demand* (Figure 2*d*). A sample is made each time EWS assessments are performed by clinical staff. The sample interval is variable, as the physician determines the rate of EWS assessments for a given patient on a daily basis. Samples are labelled sepsis positive if sepsis onset occurs within the prediction window, and sepsis negative otherwise.

Approaches *e–h* were not selected for evaluation, as the purpose of this study was not to test an exhaustive list of framing approaches, but instead to demonstrate the effect of framing on a general ward population. However, the four approaches are still relevant for reference, as they could be optimal in other clinical settings.

- *Sequential approach with prediction window* (Figure 2*e*). Data for one patient is considered a single sample with multiple prediction times. This is in contrast to the sliding window approach, where each prediction time is considered an independent sample. The observation window expands as the patient is hospitalised for a longer period of time. This is again in contrast to the sliding window approach where each independent sample has a fixed-size observation window. The prediction window is of fixed size and moves along with prediction time, generating a positive sample if sepsis occurs within the prediction window.

- *Sequential approach with entire admission as prediction window* (Figure 2*f*). Similar to the sequential approach above, except that here the problem is not predicting whether sepsis onset will occur within the moving fixed size prediction window, but if hospitalisation will lead to sepsis *at any time*.

- *At event* (Figure 2*g*). Sample generation is linked up to specific events, such as admission, preoperative assessment, or intubation.

- *Random time to onset* (Figure 2*h*). Similar to the fixed time to onset approach, except that the time of prediction for sepsis positive samples is chosen randomly at some interval before sepsis onset.

**Prediction model**

For all framing approaches, we used the extreme gradient boosting (XGBoost) [35] library, currently one of the most applied machine learning frameworks. We utilised XGBoost for all risk prediction models as this study was not an evaluation of the performance of a specific algorithm, but a study in the concept and consequences of framing a risk prediction model. XGBoost is a specific implementation of gradient tree boosting, which works by producing a risk prediction model (called a strong learner) in the form of an ensemble of weak risk prediction models (weak learners), typically decision trees. The individual risk prediction models are built sequentially by adding several decision trees, each of which is trained to minimise the risk residuals for the previous model [36,37]. Further details about implementation can be found in the linked code repository.

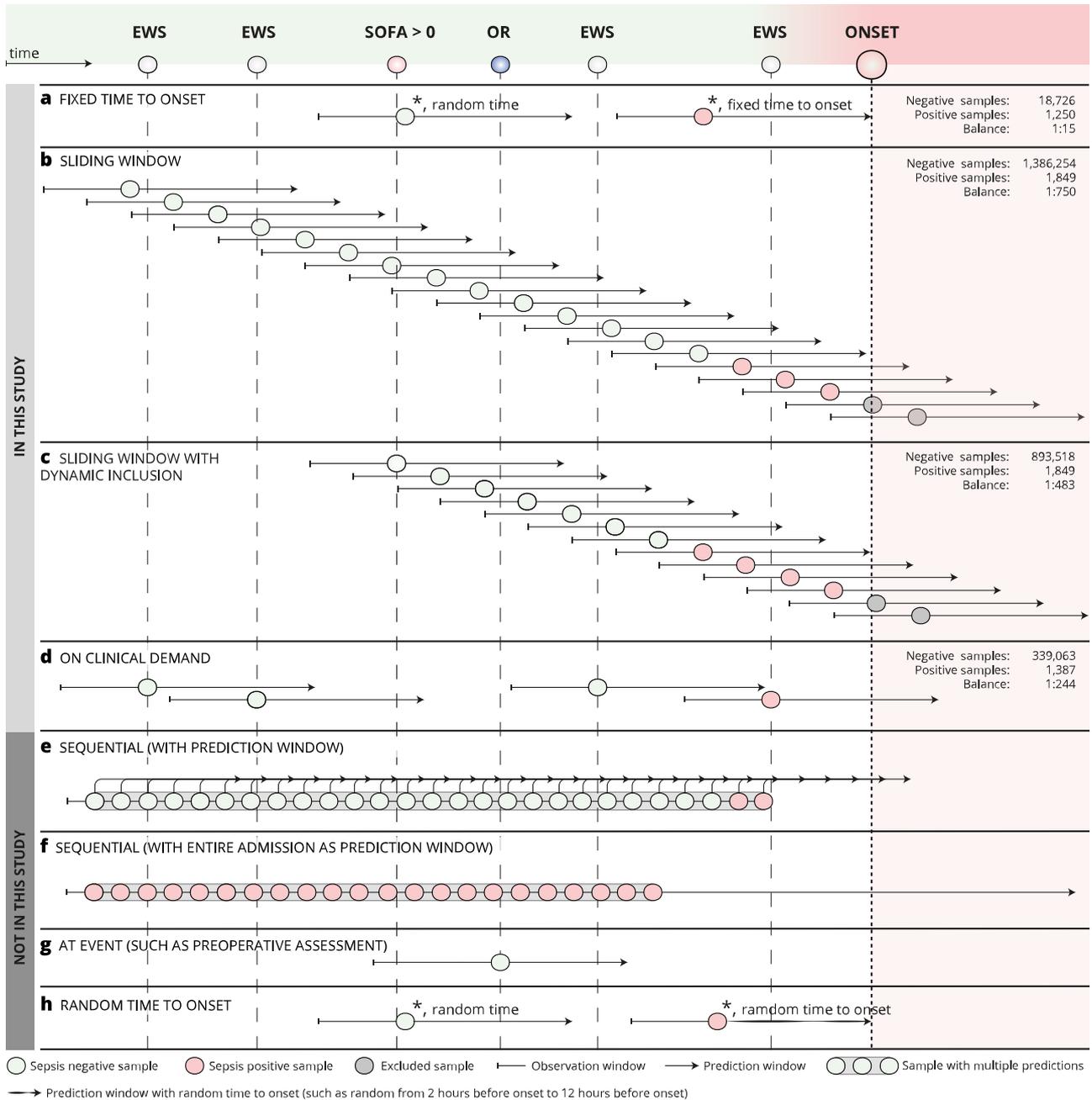

**Figure 2 | Framing of the machine learning algorithm.** Eight different approaches to frame the sepsis prediction model are shown. Fixed time to onset (**a**), sliding window (**b**), sliding window with dynamic inclusion (**c**), on clinical demand (**d**), sequential with prediction window (**e**), sequential with entire admission as prediction window (**f**), at event (**g**), and random time to onset (**h**). The top four approaches (**a–d**) are explored in this study, and the bottom four (**e–h**) are not. * indicates that the samples are from two different patients/admissions. EWS: Early Warning Score, SOFA: Sequential Organ Failure Assessment Scores, OR: Operating Room.

**Explaining predictions**

Shapley additive explanations (SHAP) [36] is a game theoretic approach to explaining the individual predictions in a risk prediction model. SHAP implements "Shapley values," which define how to fairly distribute the payout among cooperative players in a game. In SHAP, the risk prediction model is explained by declaring the parameters (i.e. covariate or confounder) "players" in a game where the prediction is the "payout." SHAP values are on the same scale as the model outcome in the risk prediction—in the example of the predictions produced for this study, this would be the probability of sepsis. A negative SHAP value indicates decreased probability of the outcome, and a positive value indicates increased probability. SHAP values sum to the difference between the predicted output for the given individual and the expected output of the model across the population. In other words, each SHAP value expresses the marginal effect that the observed parameter for the individual has on the final prediction rather than just predicting the prevalence.

One can explore each parameter's average impact on the prediction model by calculating the means of the absolute SHAP values across all individuals. Parameters with a high impact on the predicted outcome are identified by ranking these means by their magnitude (Figure 4). A plot is used to identify parameters of high importance for the prediction of the outcomes. The plot cannot show whether there are large effects of outlying parameters on a few individuals, as it only depicts the means of the absolute SHAP values.

A SHAP summary plot is a bird's-eye view of parameter importance and what is driving it. The plot is made up of many dots—one dot for each observed parameter for an individual in the investigated sample. The vertical location of the dot defines the parameter, and the horizontal position is the computed SHAP value for that parameter. The colour shows the parameter value—low values are marked in blue, and high values are marked in red. Values in the middle of the range are marked in purple. If a "swarm" of dots is centred around zero, the parameter has no effect on the model output; if not, it does.

As an example, we can consider respiratory frequency, which is the most important parameter in Figure 3d. In general, the model associates high respiratory frequency (red) with an increased of risk sepsis, and low respiratory frequency (blue) with a low risk of sepsis. An average respiratory frequency (purple) is not associated with either a high or low risk of sepsis. The vertical extent of the swarm of dots correlates with the number of measurements and clearly states that more low respiratory frequency measurements than high respiratory frequency have been measured.

A SHAP summary plot can be used to detect parameters for which outlying parameter values are important and which cannot be detected on a plot of global parameter importance; the two plots are therefore usually shown side-by-side.

**Analytic considerations**

*Data pre-processing*

In the data extracted from the CROSS-TRACKS cohort, each sample is represented using 12-hour observation windows with data for clinical parameters (19 blood tests and six vital signs, shown in Table 1). The data from the 12-hour observation window was extracted as two timesteps, each of a 6-hour duration. In the case of multiple measurements within an hour, the mean was calculated and used in place of an individual measurement. A forward and backward imputation operation was performed to lower the number of missing values. Finally, the Δ parameters (the difference between the two timesteps) were calculated to create a new set of parameters, doubling the total number of parameters to 50.

**Table 1 | List of clinical parameters**

| Laboratory parameters | | |
| --- | --- | --- |
| P(aB)-Hydrogen carbonate | P(aB)-Potassium | P-Bilirubin |
| P(aB)-pO2 | B-Leukocytes | P-Potassium |
| P(aB)-pCO2 | B-Neutrophils | P-Glucose |
| P(aB)-pH | B-Platelets | P-C-reactive protein (CRP) |
| P(aB)-Lactate | P-Sodium | Glomerular filtration rate (eGFR) |
| P(aB)-Sodium | P-Albumin | |
| P(aB)-Chloride | P-Creatinine | |

| Vital sign parameters | | |
|---|---|---|
| Systolic blood pressure | Respiratory frequency | SpO$_2$ (Pulse oximetry) |
| Diastolic blood pressure | Pulse | Temperature |

*Evaluation*

All risk prediction models were validated using five-fold cross-validation. Data were randomly divided into five portions of 20% each. For each fold, four portions (80 %) were used to fit the risk prediction model parameters during training. The remaining 20% was split into two portions of 10% each for validation and testing. The validation data were used to perform an unbiased evaluation of a model fit during training, and the test data were used to provide an unbiased evaluation of the final models. All data for each patient were assigned to either training, validation, or test data.

Figure 3 reports performance from the test data. For each fold, data were shifted such that a new portion was used for testing. The cross-validation scheme is illustrated in Supplementary Figure 1. As comparative measures for the predictive performance, we used the area under the receiver operating characteristic curve (AUROC) and the area under the precision-recall curve (AUPRC).

SHAP analysis was conducted for each fold in cross-validation, giving a total of five individual SHAP analyses. These five SHAP analyses were combined in one pooled analysis, which formed the basis for the explanatory results given in Figure 4 and Figure 5.

# Results

**Predictive performance**

In Figure 3, the predictive power of the four framing approaches is presented in summary form in terms of the AUROC and AUPRC. The AUROC with mean values and 95% confidence intervals (CIs) over the five cross-validations fold were .82 (.81–.83), .78 (.77–.79), .74 (.72–.77), and .76 (.74–.79) for fixed time to onset, sliding window, sliding window with dynamic inclusion, and on clinical demand, respectively. The AUPRC with mean values and 95% CIs were .385 (.358–.408), .007 (.006–.009), .009 (.008–.010), .014 (.011–.017) for fixed time to onset, sliding window, sliding window with dynamic inclusion, and on clinical demand, respectively.

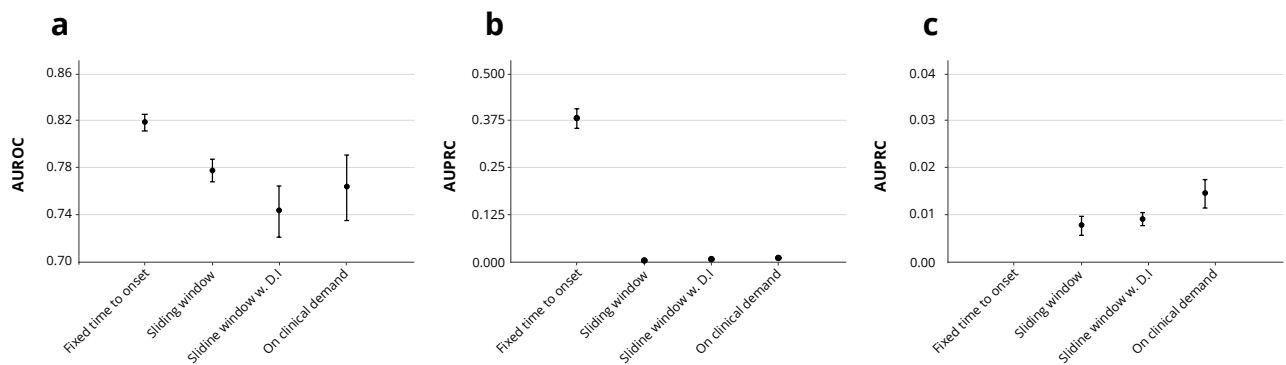

**Figure 3 | Performance evaluation.** The area under the receiver operating characteristic curve (AUROC) is shown in (**a**), and the area under the precision-recall curve (AUPRC) is shown in (**b**). In (**c**), a magnified version of (**b**) is given. Here the y-axis is bounded by AUPRC values between 0 and .04 to show details lost in (**b**) due to different scales. Error bars indicate uncertainty with 95% confidence intervals calculated from the five test datasets (n = 19,976 inpatient admissions examined over five cross-validation folds with a test size of 10%).

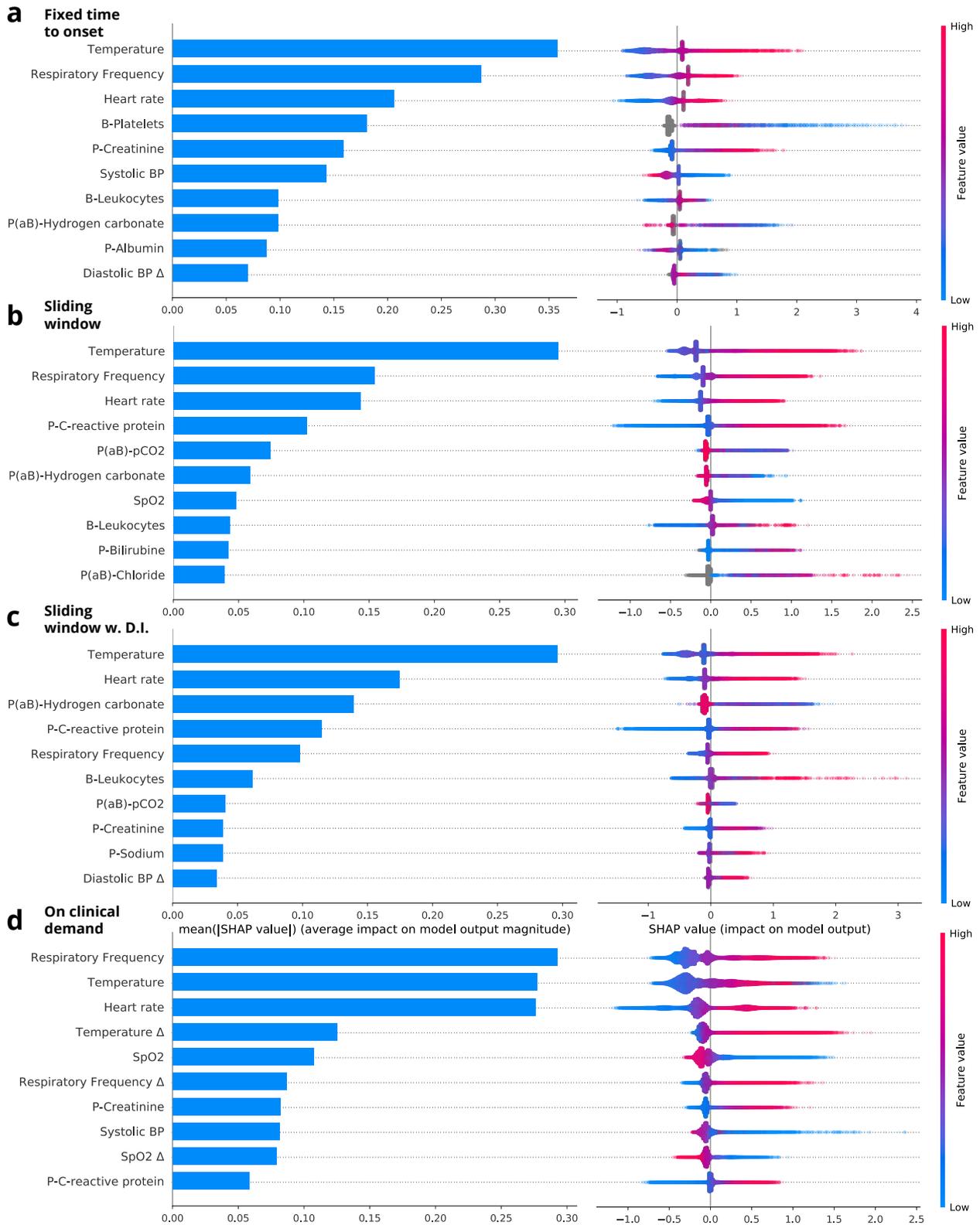

**Figure 4 | SHAP explanations.** The blue horizontal bars in the left column display the means of the absolute SHAP values. In the local explanation summary in the right-hand column, the distributions of the SHAP values for each clinical parameter are shown and colour-coded by the parameter value associated with the local explanation. Fixed time to onset is shown in **(a)**, sliding window is shown in **(b)**, sliding window with dynamic inclusion is shown in **(c)**, and on clinical demand is shown in **(d)**.

**Explanations**

In Figure 4, the ten most important clinical parameters for each model of the four framing approaches are shown. The parameters are sorted by the decreasing means of the absolute SHAP values for all individuals in the dataset. The blue horizontal bars in the left column of Figure 4 show the means of the absolute SHAP values. In the local explanation summary in the right-hand column of Figure 4, the distributions of the SHAP values for each clinical parameter are shown and colour-coded by the parameter value associated with the local explanation. Table 2 shows the percentage of missing values for the vital sign parameters for each of the four framing approaches. It can be observed that the percentage of missing values is significantly lower in the on-clinical-demand approach than in the other three approaches.

Figure 5 shows the dependence between the $SpO_2$ values and the associated $SpO_2$ SHAP values for the entire dataset. It is clear that the dependence curve is different for the fixed-time-to-onset framing approach than for the other three framing approaches, which follow a similar dependence curve. In the fixed-time-to-onset framing approach, it appears that higher $SpO_2$ values are associated with higher SHAP values and therefore have a higher probability of sepsis.

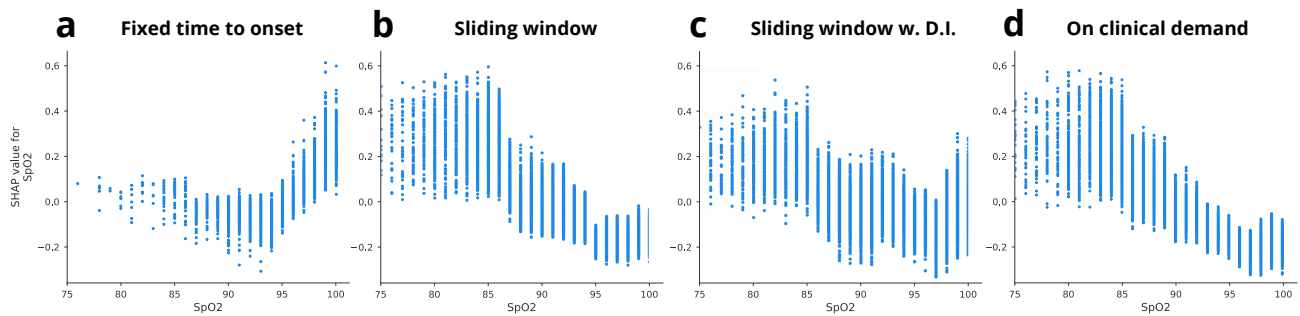

**Figure 5 | SHAP dependence plot $SpO_2$.** Fixed time to onset is shown in **(a)**, sliding window is shown in **(b)**, sliding window with dynamic inclusion is shown in **(c)**, and on clinical demand is shown in **(d)**

# Discussion

We have shown how a machine learning risk-prediction model, when applied to a dataset obtained from general medical and surgical departments, can be framed in different ways, and how this framing has direct consequences for both the model's predictive performance and the model's learning. The profound consequences of framing mandate attention from clinicians and AI developers, as understanding and reporting of framing are pivotal for the successful development and clinical implementation of future AI technology. Model framing must reflect the expected clinical environment. The importance of proper problem framing is by no means exclusive to sepsis prediction, but rather applies to most clinical risk prediction models.

In contrast to data-rich ICUs, medical and surgical departments only register clinical parameters occasionally and at random intervals, rendering real-time predictions with very high update frequencies unsuitable. It is therefore essential to look towards ways of framing the machine learning risk-prediction model in a manner optimised to the clinical environment of model deployment. The different framing approaches led to a variation in ratio between sepsis positive and sepsis negative samples between 1:15 and 1:750 (Figure 2). The derived clinical consequences of basing a model on unbalanced data could not be immediately observed using AUROC, but AUPRC varied greatly and provided a deeper understanding of the models' performance (Figure 3). The AUPRC variation is important in terms of comparing studies and methodologies and when assessing models prior to clinical use. The difference in consistency between the evaluation metrics arises inherently from their definitions. AUPRC is based on precision (*true positives* ÷ [*true positives* + *false positives*]) and recall (*true positives* ÷ [*true positives* + *false negatives*]), and AUROC is based on the true positive rate, which is equal to the recall, and the false positive rate (*false positives* ÷ [*true*

*positives + true negatives*]). Because precision-recall curves do not use true negatives at all, the AUPRC will not be affected by a large proportion of true negatives in the data, as it will focus on how the model handles the small fraction of positive examples. If the model handles the positive class well, the AUPRC will be high; if the model handles the positive class poorly, the AUPRC will be low [38].

**Table 2 | Missing values for vital sign parameters**

| Parameter | Fixed time to onset (%) | Sliding window (%) | Sliding window w. D.I (%) | On clinical demand (%) |
|---|---|---|---|---|
| Heart rate | 36.61 | 36.61 | 34.47 | 0.00 |
| Respiratory frequency | 38.85 | 38.85 | 36.81 | 0.00 |
| $SpO_2$ | 36.70 | 36.70 | 34.27 | 0.00 |
| Systolic BP | 36.45 | 36.45 | 34.29 | 0.00 |
| Diastolic BP | 36.51 | 36.51 | 34.36 | 0.07 |
| Temperature | 42.62 | 42.62 | 40.52 | 5.88 |
| Heart rate Δ | 80.86 | 80.86 | 78.30 | 26.81 |
| Respiratory frequency Δ | 81.42 | 81.42 | 78.96 | 26.81 |
| $SpO_2$ Δ | 80.91 | 80.91 | 78.30 | 26.81 |
| Systolic BP Δ | 80.79 | 80.79 | 78.21 | 26.81 |
| Diastolic BP Δ | 80.80 | 80.80 | 78.22 | 26.85 |
| Temperature Δ | 82.42 | 82.42 | 79.99 | 30.95 |

The on-clinical-demand framing, which involved samples for each time clinicians made an EWS assessment in the clinic, showed the lowest percentage of missing values among the vital sign parameters (Table 2); this model was also able to learn more temporal dependencies than the others.

In Figure 5, the SHAP values are plotted against $SpO_2$ values in a SHAP dependence plot. In these plots, it can be seen that the models have each learned to interpret $SpO_2$ differently. The fixed-time-to-onset model associates high $SpO_2$ with the development of sepsis, which is, of course, wrong. This unfortunate association might come from biases introduced in the framing and emphasises the importance of both the right framing as well as the need for transparent risk-prediction models that can be verified by clinicians [23].

**Conceptualisation of framing in machine learning algorithms in published literature**

Details on framing in previous published sepsis risk prediction models are difficult to obtain (summarised in Table 3). In the case of sepsis risk prediction models, framing is typically left or right aligned [26]. Left-aligned models predict the onset of sepsis at fixed events during admission, such as on admission or preoperatively. Right-aligned models continuously predict whether sepsis will occur in some prediction window and are often named as real-time or continuous-prediction models. Most studies report on right-aligned models, but there are great variations to problem framing even within the right-aligned category.

Barton et al. and Lauritsen et. al [15,23] both used fixed-time-to-onset with fixed-distance, look-ahead windows. In both studies, patients who never developed sepsis were assigned an onset time at random according to a continuous, uniform probability distribution. Scherpf et al. and Moor et al.[19,22] both framed the sepsis prediction problem as a sequential problem that was solved with long short-term memory (LSTM)[39] and temporal convolutional (TCN) networks [40,41]. Futoma et al. also framed the prediction problem as a sequential problem. Given a new patient encounter, their goal was to continuously update the predicted probability that an encounter would result in sepsis, at any time during admission [18]. As such, they did not predict whether sepsis would occur within a fixed prediction window, but instead if the admission would result in sepsis, at any time. Wyk et al. used a sliding-window approach, in which samples were generated by sliding the observation and prediction windows with small shifts in time [7]. Nemati used a similar sliding-window approach [10] and predicted for each shift in time if sepsis onset occurred within a fixed-prediction window—in contrast to the work of Futoma et al. In a later work by Futoma et al., the authors employed a more sophisticated and upgraded version of the fixed-time-to-onset approach [9]. Khojandi et al. used both [42] left- and right-aligned models. For their left-aligned model, they predicted the risk of sepsis at 12, 24, and 48 hours after admission. For their right-aligned model, the authors predicted sepsis onset 12, 24, and 48 hours before onset with random sampled negative onset times. Khoshnevisan et al. used fixed time to onset, but for the negative cases, instead of uniform sampling, the authors used the last-registered timestamp in the admission as the onset time [43].

**Table 3 | Framing approaches in related studies**

| Paper | Population | Target | Problem framing | Alignment |
|---|---|---|---|---|
| Van Wyk, 2018 [7] | Intensive care unit | Sepsis | Sliding window | Right |
| Scherpf, 2019 [22] | Intensive care unit | Sepsis | Sequential problem | Right |
| Moor, 2019 [19] | Intensive care unit | Sepsis | Sequential problem | Right |
| Futoma I, 2017 [18] | Mixed-ward | Sepsis | Sequential problem (modified) | Right |
| Futoma II, 2017 [9] | Mixed-ward | Sepsis | Fixed time to onset + matching | Right |
| Lauritsen I, 2020 [12] | Mixed-ward | Sepsis | Sequential problem | Right |
| Lauritsen II, 2020 [23] | Mixed-ward | Sepsis | Fixed time to onset | Right |
| Nemati, 2018 [10] | Intensive care unit | Sepsis | Sliding window | Right |
| Delahanty, 2019 [44] | Emergency department | Sepsis | On-clinical demand (modified) | Right |
| Khojandi, 2018 [42] | In-hospital | Sepsis | Fixed time to onset, fixed time | Left & right |
| Khoshnevisan, 2018 [43] | In-hospital | Septic shock | Fixed time to onset (modified) | Left & right |
| Thiel, 2019 [45] | In-hospital | Septic shock | Fixed time to onset (modified) | Right |
| Van Wyk I, 2018 [46] | Intensive care unit | Sepsis | Sliding window | Right |
| Wang, 2018 [47] | Intensive care unit | Sepsis | Fixed time to onset | Right |
| Kam, 2017 [24] | Intensive care unit | Sepsis | Sliding window | Right |
| Moss, 2016 [48] | Intensive care unit | Severe sepsis | Sliding window | Right |
| Guillen, 2015 [49] | Intensive care unit | Severe sepsis | Fixed time to onset | Right |
| Mao, 2017 [16] | Mixed-ward | Sepsis, severe sepsis, septic shock | Fixed time to onset | Right |
| Barton, 2019 [15] | Mixed-ward | Sepsis | Fixed time to onset | Right |

Given the substantial variation of framing approaches in previously published risk prediction models, it may not be wise to compare the derived results directly. Instead, we encourage a greater focus on developing and sharing datasets that are already framed, in order to provide a much better basis for comparison. One example of such a dataset developed to address a specific clinical problem is Early Prediction of Sepsis from Clinical Data: The PhysioNet/Computing in Cardiology Challenge 2019, which is based on the MIMIC-III database [50]. This dataset comes with a utility function that rewards early predictions and penalises late predictions as well as false alarms in a consistent way.

**Strength and limitations**

A major strength of this study is that we created a language for discussing the concept of framing as essential to building machine learning risk-prediction models. On a large, population-based, open cohort we showed that the same machine learning architecture, applied to the same data, gave rise to many different models when framing is varied. We believe inclusion of this aspect in future risk-prediction models is fundamental to enabling healthcare professionals to discuss the clinical value of developed models or models under development with technical professionals who construct the machine learning models.

The study is not without limitations. It is a weakness that we did not test multiple models, such as sequential models, as they are widely applied in sepsis prediction literature. However, it is imperative to stress that even if we had included sequential models, they would still need to be framed and evaluated at specific times, exactly as it was done for the chosen models in this study. Also, XGBoost does not support sequential modelling, thus it would be necessary to introduce a second risk-prediction model with manual imputation of missing values.

We defined "suspected infection" in accordance with previous studies in the field, but the chosen approach may have led to the misclassification of some patients into the wrong sepsis group (positive/negative). In terms of the SHAP analysis, the addition of sequential models would lead to explanations over multiple timesteps that would have to be equated.

Another limitation is that this study only concerned sepsis risk prediction in secondary care, and findings may vary in different settings and for other outcomes. However, the fundamentals of how framing is directly linked to the clinical problem, which the machine learning model must solve, do not change.

# Conclusion

Problem framing is crucial, as all the subsequent work of training and evaluating will be in the context of how the problem has been framed, and explicit documentation thereof will make it easier to compare evaluation metrics between published studies. We therefore recommend that problem framing should be made very clear in future articles.

**Contributors:** S.M.L, J.L, M.J.J., and B.T. initiated the project. B.T., S.M.L., J.L, J.B.W., U.S.E., and M.J.J. contributed to the overall experimental design. S.M.L created the dataset. S.M.L. contributed to the software engineering and together with B.T., S.M.L., and A.H.R analysed the results. S.M.L. created the first draft of the article. All authors contributed significantly to revision of the first draft of the article and approved of the final version of the manuscript.

**Funding:** This work was funded by the Innovation Fund Denmark (case number 8053-00076B).

**Competing interests:** All authors have completed the ICMJE uniform disclosure form at www.icmje.org/coi_disclosure.pdf and declare that S.M.L, A.H.R, and B.T. are employed at Enversion. The authors have no other competing interests to disclose.

**Ethics approval:** The study was approved by the Danish Data Protection Agency (case number 1-16-02-541-15). Additionally, the data used in this work were collected with the approval of the steering committee for CROSS-TRACKS. Only retrospective data were used for this research, without the active involvement of patients or potential influence on their treatment. Therefore, under current national laws, no formal ethical approval was necessary.

**Data sharing:** The authors have accessed the data referred to herein through the CROSS-TRACKS cohort, which is a newer Danish cohort that combines primary and secondary sector data. Due to EU regulations, specifically the GDPR, these data are not readily available to the wider research community per se. However, all researchers can apply for access to the data by following the instructions on this page: http://www.tvaerspor.dk/.
   The code used to conduct the analyses is shared here: https://github.com/SimonMeyerLauritsen/Framing. For data extraction, our experimental framework for data makes use of proprietary libraries that belong to Enversion A/S, and we are unable to publicly release this code. We have described the experiments and implementation details in the Methods section to allow for independent replication. Further inquiry regarding the specific nature of the AI model can be made by relevant parties to the corresponding author.

**Transparency declaration:** The lead author affirms that the manuscript is an honest, accurate, and transparent account of the study being reported, that no important aspects of the study have been omitted, and that any discrepancies from the study as planned (and, if relevant, registered) have been explained.

**Dissemination to participants and related patient and public communities:** Dissemination to research participants is not relevant.

**Acknowledgements:** We acknowledge the steering committee for CROSS-TRACKS for access to the data. For help with acquisition, modelling, and validation of the data extraction pipelines, we thank Per Dahl Rasmussen, Rasmus Holm Laursen, Anne Olsvig Boilesen, Emil Møller Bartels, and Christian Bang. We also thank the rest of the Enversion team for their support.

# References

1 Topol EJ. High-performance medicine: the convergence of human and artificial intelligence. *Nat Med* 2019;25:44 56. Available from: doi:10.1038/s41591-018-0300-7

2 Kelly CJ, Karthikesalingam A, Suleyman M, *et al.* Key challenges for delivering clinical impact with artificial intelligence. *Bmc Med* 2019;17:195. Available from: doi:10.1186/s12916-019-1426-2

3 Cosgriff CV, Stone DJ, Weissman G, *et al.* The clinical artificial intelligence department: a prerequisite for success. *Bmj Heal Care Informatics* 2020;27:e100183. Available from: doi:10.1136/bmjhci-2020-100183

4 Higgins D, Madai VI. From bit to bedside: A practical framework for artificial intelligence product development in healthcare. *Adv Intelligent Syst* 2020;2:2000052. Available from: doi:10.1002/aisy.202000052

5 Li Y, Sperrin M, Ashcroft DM, *et al.* Consistency of variety of machine learning and statistical models in predicting clinical risks of individual patients: longitudinal cohort study using cardiovascular disease as exemplar. *Bmj* 2020;371:m3919. Available from: doi:10.1136/bmj.m3919

6 Calvert JS, Price DA, Chettipally UK, *et al.* A computational approach to early sepsis detection. *Comput Biol Med* 2016;74:69 73. Available from: doi:10.1016/j.compbiomed.2016.05.003

7 Wyk F van, Khojandi A, Mohammed A, *et al.* A minimal set of physiomarkers in high frequency real-time physiological data streams predict adult sepsis onset earlier. *Int J Med Inform* 2018;122:55–62. Available from: doi:10.1016/j.ijmedinf.2018.12.002

8 Kaji DA, Zech JR, Kim JS, *et al.* An attention based deep learning model of clinical events in the intensive care unit. *Plos One* 2019; Available from: 14:e0211057. doi:10.1371/journal.pone.0211057

9 Futoma J, Hariharan S, Sendak M, *et al.* An improved multi-output gaussian process rnn with real-time validation for early sepsis detection. *Proceedings of Machine Learning for Healthcare 2017*; 2017.

10 Nemati S, Holder A, Razmi F, *et al.* An interpretable machine learning model for accurate prediction of sepsis in the ICU. *Crit Care Med* 2017;46:547–53. Available from: doi:10.1097/ccm.0000000000002936

11 Liu R, Greenstein JL, Granite SJ, *et al.* Data-driven discovery of a novel sepsis pre-shock state predicts impending septic shock in the ICU. *Sci Rep-uk* 2019;9:6145. Available from: doi:10.1038/s41598-019-42637-5

12 Lauritsen SM, Kalør ME, Kongsgaard EL, *et al.* Early detection of sepsis utilizing deep learning on electronic health record event sequences. *Artif Intell Med* 2020;101820. Available from: doi:10.1016/j.artmed.2020.101820

13 Shimabukuro DW, Barton CW, Feldman MD, *et al.* Effect of a machine learning-based severe sepsis prediction algorithm on patient survival and hospital length of stay: a randomised clinical trial. *Bmj Open Respir Res* 2017;4:e000234 8. Available from: doi:10.1136/bmjresp-2017-000234

14 Burdick H, Pino E, Gabel-Comeau D, *et al.* Evaluating a sepsis prediction machine learning algorithm in the emergency department and intensive care unit: a before and after comparative study. 2018. Available from: doi:10.1101/224014

15 Barton C, Chettipally U, Zhou Y, *et al.* Evaluation of a machine learning algorithm for up to 48-hour advance prediction of sepsis using six vital signs. *Comput Biol Med* 2019;109:79 84. Available from: doi:10.1016/j.compbiomed.2019.04.027

16 Mao Q, Jay M, Hoffman JL, *et al.* Multicentre validation of a sepsis prediction algorithm using only vital sign data in the emergency department, general ward and ICU. *Bmj Open* 2018;8:e017833 11. Available from: doi:10.1136/bmjopen-2017-017833

17 Vellido A, Ribas V, Morales C, *et al.* Machine Learning for Critical Care: An Overview and a Sepsis Case Study. *Springer.* Bioinformatics and Biomedical Engineering 2017 (eds Rojas, I. & Ortuño, F.) (IWBBIO) vol. 10208, 15–30 (2017). Available from:  doi:10.1007/978-3-319-56148-6_2

18 Futoma J, Hariharan S, Heller K. Learning to Detect Sepsis with a Multitask Gaussian Process RNN Classifier. In: Proceedings of the 34th International Conference on Machine Learning. ICML vol. 70, 1174–1182 (2017).

19 Moor M, Horn M, Rieck B, *et al.* Temporal convolutional networks and dynamic time warping can drastically improve the early prediction of sepsis. In: Proceedings of the 4th Machine Learning for Healthcare Conference. PMLR 106 (2019).

20 Choi E, Bahadori MT, Kulas JA, *et al.* RETAIN: An interpretable predictive model for healthcare using reverse time attention mechanism. In 30th Annual Conference on Neural Information Processing Systems (NIPS 2016). Advances in Neural Information Processing Systems3512–3520 (2016).

21 Islam MM, Nasrin T, Walther BA, *et al.* Prediction of sepsis patients using machine learning approach: A meta-analysis. *Comput Meth Prog Bio* 2019;170:1 9. Available from:  doi:10.1016/j.cmpb.2018.12.027

22 Scherpf M, Gräßer F, Malberg H, *et al.* Predicting sepsis with a recurrent neural network using the MIMIC III database. *Comput Biol Med* 2019;113:103395. Available from: doi:10.1016/j.compbiomed.2019.103395

23 Lauritsen SM, Kristensen M, Olsen MV, *et al.* Explainable artificial intelligence model to predict acute critical illness from electronic health records. *Nat Commun* 2020;11:3852. Available from: doi:10.1038/s41467-020-17431-x

24 Kam HJ, Kim HY. Learning representations for the early detection of sepsis with deep neural networks. *Comput Biol Med* 2017;89:248–55. Available from: doi:10.1016/j.compbiomed.2017.08.015

25 Shickel B, Loftus TJ, Adhikari L, *et al.* DeepSOFA: A continuous acuity score for critically ill patients using clinically interpretable deep learning. *Sci Rep-uk* 2019;9:1879. Available from: doi:10.1038/s41598-019-38491-0

26 Fleuren LM, Klausch TLT, Zwager CL, *et al.* Machine learning for the prediction of sepsis: a systematic review and meta-analysis of diagnostic test accuracy. *Intens Care Med* 2020;46:383–400. Available from: doi:10.1007/s00134-019-05872-y

27 Johnson AEW, Pollard TJ, Shen L, *et al.* MIMIC-III, a freely accessible critical care database. *Sci Data* 2016;3:160035 9. Available from: doi:10.1038/sdata.2016.35

28 Johnson AEW, Stone DJ, Celi LA, *et al.* The MIMIC code repository: enabling reproducibility in critical care research. *J Am Med Inform Assn* 2017;25:32–9. Available from: doi:10.1093/jamia/ocx084

29 Sakr Y, Jaschinski U, Wittebole X, *et al.* Sepsis in intensive care unit patients: worldwide data from the intensive care over nations audit. *Open Forum Infect Dis* 2018;5:ofy313. Available from: doi:10.1093/ofid/ofy313

30 Riis AH, Kristensen PK, Petersen MG, *et al.* Cohort profile: CROSS-TRACKS: a population-based open cohort across healthcare sectors in Denmark. *Bmj Open* 2020;10:e039996. Available from: doi:10.1136/bmjopen-2020-039996

31 Pedersen CB. The danish civil registration system. *Scand J Public Healt* 2011;39:22–5. Available from: doi:10.1177/1403494810387965

32 Seymour CW, Liu VX, Iwashyna TJ, *et al.* Assessment of clinical criteria for sepsis: for the third international consensus definitions for sepsis and septic shock (sepsis-3). *Jama* 2016;315:762–74. Available from: doi:10.1001/jama.2016.0288

33 Singer M, Deutschman CS, Seymour CW, *et al.* The third international consensus definitions for sepsis and septic shock (sepsis-3). *Jama* 2016;315:801 10. Available from: doi:10.1001/jama.2016.0287

34 Vincent J-L, Moreno R, Takala J, *et al.* The SOFA (sepsis-related organ failure assessment) score to describe organ dysfunction/failure. *Intens Care Med* 2018;22:707–10. Available from: doi:10.1007/bf01709751

35 Chen T, Guestrin C. XGBoost: A scalable tree boosting system. 2016;785–94. Available from: doi:10.1145/2939672.2939785

36 Lundberg SM, Erion G, Chen H, *et al.* From local explanations to global understanding with explainable AI for trees. *Nat Mach Intell* 2020;2:56–67. Available from: doi:10.1038/s42256-019-0138-9

37 Lundberg SM, Lee S-I. A unified approach to interpreting model predictions. 2017. 4765–4774. Available from: http://papers.nips.cc/paper/7062-a-unified-approach-to-interpreting-model-predictions.pdf

38 Saito T, Rehmsmeier M. The precision-recall plot is more informative than the roc plot when evaluating binary classifiers on imbalanced datasets. *Plos One* 2015;10:e0118432. Available from: doi:10.1371/journal.pone.0118432

39 LeCun Y, Bengio Y, Hinton G. Deep learning. *Nature* 2015;521:436–444. Available from: doi:10.1038/nature14539

40 Lea C, Vidal R, Reiter A, *et al.* Temporal convolutional networks: a unified approach to action segmentation. In Computer Vision – ECCV 2016 Workshops. 47–54 (Springer International Publishing, 2016).

41 Bai S, Kolter JZ, Koltun V. An empirical evaluation of generic convolutional and recurrent networks for sequence modeling. Preprint at https://arxiv.org/abs/1803.01271 (2018).

42 Khojandi A, Tansakul V, Li X, *et al.* Prediction of sepsis and in-hospital mortality using electronic health records. *Method Inform Med* 2018;57:185–93. Available from: doi:10.3414/me18-01-0014

43 Khoshnevisan F, Ivy J, Capan M, *et al.* Recent temporal pattern mining for septic shock early prediction. *2018 Ieee Int Conf Healthc Informatics Ichi* 2018;229–40. Available from: doi:10.1109/ichi.2018.00033

44 Delahanty RJ, Alvarez J, Flynn LM, *et al.* Development and evaluation of a machine learning model for the early identification of patients at risk for sepsis. *Ann Emerg Med* 2019;73:334–44. Available from: doi:10.1016/j.annemergmed.2018.11.036

45 Schamoni S, Lindner HA, Schneider-Lindner V, *et al.* Leveraging implicit expert knowledge for non-circular machine learning in sepsis prediction. *Artif Intell Med* 2019;100:101725. Available from: doi:10.1016/j.artmed.2019.101725

46 Wyk F van, Khojandi A, Kamaleswaran R. Improving prediction performance using hierarchical analysis of real-time data: a sepsis case study. *Ieee J Biomed Health* 2018;23:978–86. Available from: doi:10.1109/jbhi.2019.2894570

47 Wang RZ, Sun CH, Schroeder PH, *et al.* Predictive models of sepsis in adult ICU patients. *2018 Ieee Int Conf Healthc Informatics Ichi* 2018;390–1. Available from: doi:10.1109/ichi.2018.00068

48 Moss TJ, Lake DE, Calland JF, *et al.* Signatures of subacute potentially catastrophic illness in the ICU. *Crit Care Med* 2016;44:1639–48. Available from: doi:10.1097/ccm.0000000000001738

49 Guillon J, Liu J, Furr M, *et al.* Predictive models for severe sepsis in adult ICU patients. *2015 Syst Information Eng Des Symposium* 2015;:182–7. Available from: doi:10.1109/sieds.2015.7116970

50 Reyna MA, Josef C, Seyedi S, *et al.* Early prediction of sepsis from clinical data: the PhysioNet/computing in cardiology challenge 2019. *Comput Cardiol Cinc* 2019;00:1–4. Available from: doi:10.23919/cinc49843.2019.9005736

SUPPLEMENTARY INFORMATION FOR

# The Consequences of the *Framing* of Machine Learning Risk Prediction Models: Evaluation of Sepsis in General Wards

**Simon Meyer Lauritsen**[1,2], **Bo Thiesson**[1,3], **Marianne Johansson Jørgensen**[4], **Anders Hammerich Riis**[1], Ulrick Skipper Espelund[4,5], Jesper Bo Weile[6,7], Jeppe Lange[2,4]

[1]*Enversion A/S, Fiskerivej 12, 1st floor, 8000 Aarhus C, Denmark;* [2]*Department of Clinical Medicine, Aarhus University, Aarhus N, Denmark;* [3]*Department of Engineering, Aarhus University, Aarhus C, Denmark;* [4]*Department of Research, Horsens Regional Hospital, Horsens, Denmark; Department of Anesthesiology, Horsens Regional Hospital, Denmark;* [6]*Emergency Department, Horsens Regional Hospital, Denmark;* [7]*Research Center for Emergency Medicine, Aarhus University Hospital, Denmark*

*\*Corresponding author: Simon Meyer Lauritsen (sla@enversion.dk)*

**Supplementary table 1 – missing values for parameters used in the four models**

| Parameter | On clinical demand (%) | Sliding window (%) | Fixed time to onset (%) | Sliding window w. dynamic incl. (%) |
|---|---|---|---|---|
| Heart rate | 0.00 | 36.61 | 24.97 | 34.47 |
| Respiratory Frequency | 0.00 | 38.85 | 26.65 | 36.81 |
| SpO2 | 0.00 | 36.70 | 25.06 | 34.27 |
| Systolic BP | 0.00 | 36.45 | 24.77 | 34.29 |
| Diastolic BP | 0.07 | 36.51 | 24.88 | 34.36 |
| Temperature | 5.88 | 42.62 | 31.59 | 40.52 |
| Heart rate Δ | 26.81 | 80.86 | 71.48 | 78.30 |
| Respiratory Frequency Δ | 26.81 | 81.42 | 72.03 | 78.96 |
| SpO2 Δ | 26.81 | 80.91 | 71.53 | 78.30 |
| Systolic BP Δ | 26.81 | 80.79 | 71.33 | 78.21 |
| Diastolic BP Δ | 26.85 | 80.80 | 71.37 | 78.22 |
| Temperature Δ | 30.95 | 82.42 | 73.71 | 79.99 |
| P-Sodium | 71.77 | 69.95 | 60.99 | 66.54 |
| P-Potassium | 71.84 | 70.01 | 61.14 | 66.63 |
| P-Creatinine | 71.96 | 70.12 | 61.39 | 66.72 |
| P-Albumin | 73.47 | 70.95 | 62.27 | 67.65 |
| B-Leukocytes | 73.59 | 72.27 | 62.14 | 69.49 |
| P-C-reactive protein | 73.92 | 72.78 | 63.71 | 69.92 |
| P-Sodium Δ | 75.90 | 89.21 | 84.55 | 87.01 |
| P-Potassium Δ | 75.93 | 89.23 | 84.62 | 87.03 |
| P-Creatinine Δ | 76.05 | 89.26 | 84.74 | 87.07 |
| P-Albumin Δ | 77.27 | 89.63 | 85.07 | 87.51 |
| P-C-reactive protein Δ | 77.71 | 90.36 | 85.87 | 88.42 |
| B-Leukocytes Δ | 78.20 | 90.29 | 84.83 | 88.37 |
| eGFR | 81.39 | 79.56 | 74.72 | 74.70 |
| P-Glucose | 82.17 | 79.15 | 83.14 | 75.21 |
| P-Glucose Δ | 83.33 | 88.69 | 89.45 | 86.19 |
| eGFR Δ | 84.11 | 92.33 | 89.45 | 89.99 |
| B-Platelets | 90.28 | 95.00 | 93.93 | 94.09 |
| P-Bilirubine | 90.85 | 94.88 | 93.20 | 93.97 |
| P(aB)-pH | 92.15 | 93.91 | 94.65 | 91.67 |
| P(aB)-pCO2 | 92.15 | 93.91 | 94.64 | 91.67 |
| P(aB)-pO2 | 92.16 | 93.93 | 94.65 | 91.70 |
| P(aB)-Lactate | 92.18 | 93.94 | 94.67 | 91.72 |
| P(aB)-Hydrogen carbonate | 92.22 | 93.93 | 94.66 | 91.71 |
| P(aB)-Potassium | 92.35 | 94.07 | 94.75 | 91.90 |
| P(aB)-Sodium | 92.36 | 94.08 | 94.76 | 91.91 |
| B-Platelets Δ | 92.86 | 97.39 | 96.79 | 96.73 |
| P-Bilirubine Δ | 92.88 | 97.71 | 96.84 | 97.14 |
| P(aB)-pCO2 Δ | 93.59 | 95.51 | 95.79 | 93.83 |
| P(aB)-pH Δ | 93.59 | 95.50 | 95.80 | 93.83 |
| P(aB)-pO2 Δ | 93.61 | 95.52 | 95.80 | 93.84 |
| P(aB)-Lactate Δ | 93.62 | 95.52 | 95.81 | 93.85 |
| P(aB)-Hydrogen carbonate Δ | 93.65 | 95.52 | 95.80 | 93.85 |
| P(aB)-Potassium Δ | 93.78 | 95.60 | 95.87 | 93.96 |
| P(aB)-Sodium Δ | 93.79 | 95.60 | 95.88 | 93.97 |
| P(aB)-Chloride | 97.98 | 98.64 | 98.55 | 98.20 |
| P(aB)-Chloride Δ | 98.46 | 98.96 | 98.84 | 98.62 |
| B-Neutrophils | 99.98 | 99.99 | 99.96 | 99.98 |
| B-Neutrophils Δ | 99.98 | 99.99 | 99.97 | 99.99 |

**Supplementary table 2 – cohort characteristic**

| Characteristic | Sepsis negative | Sepsis positive | Total |
|---|---|---|---|
| **Number of patients** | 18,726 (94) | 1,250 (6) | 19,976 (100) |
| **Age** | | | |
| Age 0-17 | 77 (0) | 0 (0) | 77 (0) |
| Age 18-39 | 1,460 (8) | 55 (4) | 1,515 (8) |
| Age 40-64 | 4,347 (23) | 283 (23) | 4,630 (23) |
| Age 65-79 | 6,921 (37) | 522 (42) | 7,443 (37) |
| Age 80+ | 5,921 (32) | 390 (31) | 6,311 (32) |
| **Age Non-missing N (%)** | 18,726 (100) | 1,250 (100) | 19,976 (100) |
| **Age Mean (SD)** | 70 (17) | 71 (15) | 70 (17) |
| **Age Median (IQR)** | 73 (61-83) | 73 (64-82) | 73 (61-82) |
| **Gender** | | | |
| Female | 9,844 (53) | 488 (39) | 10,332 (52) |
| Male | 8,882 (47) | 762 (61) | 9,644 (48) |
| **Municipality** | | | |
| Hedensted | 3,956 (21) | 277 (22) | 4,233 (21) |
| Horsens | 9,098 (49) | 590 (47) | 9,688 (48) |
| Odder | 2,014 (11) | 133 (11) | 2,147 (11) |
| Other | 37 (0) | 1 (0) | 38 (0) |
| Skanderborg | 3,621 (19) | 249 (20) | 3,870 (19) |
| **Marital status** | | | |
| Divorced or dissolved partnership | 2,644 (14) | 187 (15) | 2,831 (14) |
| Married or registered partnership | 8,070 (43) | 577 (46) | 8,647 (43) |
| Not married | 1,721 (9) | 89 (7) | 1,810 (9) |
| Unknown | 1,122 (6) | 87 (7) | 1,209 (6) |
| Widowed or longest living in a registered partnership | 5,169 (28) | 310 (25) | 5,479 (27) |
| **Cohabitation status: Living alone** | 7,397 (40) | 499 (40) | 7,896 (40) |
| **Socioeconomic status (within 1 year prior to hospitalisation)** | | | |
| Health-related benefit | 2,799 (15) | 186 (15) | 2,985 (15) |
| Labour-market-related benefit | 514 (3) | 31 (2) | 545 (3) |
| Normal retirement | 4,323 (23) | 321 (26) | 4,644 (23) |
| Self-supporting | 11,090 (59) | 712 (57) | 11,802 (59) |
| **Health services from the municipalities (within 30 days prior to hospitalisation)** | | | |
| Rehabilitation (eg physiotherapy) | | | |
| No | 15,394 (82) | 1,038 (83) | 16,432 (82) |
| Yes | 3,332 (18) | 212 (17) | 3,544 (18) |
| Practical help (eg home cleaning) | | | |
| No | 11,984 (64) | 826 (66) | 12,810 (64) |
| Yes | 6,742 (36) | 424 (34) | 7,166 (36) |
| Personal care (eg weekly bath) | | | |
| No | 11,413 (61) | 784 (63) | 12,197 (61) |
| Yes | 7,313 (39) | 466 (37) | 7,779 (39) |
| Home visits by a community nurse (eg injection of medicine) | | | |
| No | 10,433 (56) | 697 (56) | 11,130 (56) |
| Yes | 8,293 (44) | 553 (44) | 8,846 (44) |
| Any service | | | |
| No | 8,525 (46) | 566 (45) | 9,091 (46) |
| Yes | 10,201 (54) | 684 (55) | 10,885 (54) |
| **Health services from the municipalities (within 30 days prior to hospitalisation)** | | | |
| 1 Home visits by a community nurse | 8,293 (44) | 553 (44) | 8,846 (44) |
| 2 Personal care | 1,008 (5) | 68 (5) | 1,076 (5) |
| 3 Practical help | 615 (3) | 40 (3) | 655 (3) |
| 4 Rehabilitation | 285 (2) | 23 (2) | 308 (2) |
| 5 No service | 8,525 (46) | 566 (45) | 9,091 (46) |
| **Health services from the municipalities (within 90 days prior to hospitalisation)** | | | |
| 1 Home visits by a community nurse | 8,625 (46) | 582 (47) | 9,207 (46) |
| 2 Personal care | 1,004 (5) | 64 (5) | 1,068 (5) |
| 3 Practical help | 625 (3) | 42 (3) | 667 (3) |
| 4 Rehabilitation | 356 (2) | 29 (2) | 385 (2) |
| 5 No service | 8,116 (43) | 533 (43) | 8,649 (43) |
| **Primary healthcare services (within 30 days prior to hospitalisation)** | | | |
| Daytime face-to-face contact | | | |
| No | 5,286 (28) | 358 (29) | 5,644 (28) |
| Yes | 13,440 (72) | 892 (71) | 14,332 (72) |
| Out-of-hours face-to-face contact | | | |
| No | 16,500 (88) | 1,115 (89) | 17,615 (88) |
| Yes | 2,226 (12) | 135 (11) | 2,361 (12) |
| Any GP contact | | | |
| No | 1,283 (7) | 88 (7) | 1,371 (7) |
| Yes | 17,443 (93) | 1,162 (93) | 18,605 (93) |
| Dentist (within 1 year) | | | |
| No | 9,240 (49) | 646 (52) | 9,886 (49) |
| Yes | 9,486 (51) | 604 (48) | 10,090 (51) |
| **Known in the primary setting (within 30 days prior to hospitalisation)** | | | |
| No | 1,066 (6) | 69 (6) | 1,135 (6) |
| Yes | 17,660 (94) | 1,181 (94) | 18,841 (94) |
| **Primary healthcare services (within 90 days prior to hospitalisation)** | | | |
| Daytime face-to-face contact | | | |
| No | 2,448 (13) | 166 (13) | 2,614 (13) |
| Yes | 16,278 (87) | 1,084 (87) | 17,362 (87) |
| Out-of-hours face-to-face contact | | | |
| No | 15,850 (85) | 1,073 (86) | 16,923 (85) |
| Yes | 2,876 (15) | 177 (14) | 3,053 (15) |
| Any GP contact | | | |
| No | 464 (2) | 29 (2) | 493 (2) |
| Yes | 18,262 (98) | 1,221 (98) | 19,483 (98) |
| **Known in the primary setting (within 90 days prior to hospitalisation)** | | | |
| No | 394 (2) | 25 (2) | 419 (2) |
| Yes | 18,332 (98) | 1,225 (98) | 19,557 (98) |
| **Charlson comorbidity index score (within 10 years prior to hospitalisation)** | | | |
| Score 0: No comorbidity | 8,142 (43) | 467 (37) | 8,609 (43) |
| Score 1: Low comorbidity | 4,032 (22) | 261 (21) | 4,293 (21) |
| Score 2: Medium comorbidity | 2,667 (14) | 218 (17) | 2,885 (14) |
| Score 3+ High comorbidity | 3,885 (21) | 304 (24) | 4,189 (21) |
| **Psychiatric disease (within 10 years prior to hospitalisation)** | | | |
| Mild psychiatric disease | 4,042 (22) | 262 (21) | 4,304 (22) |
| No disease | 14,219 (76) | 957 (77) | 15,176 (76) |
| Severe psychiatric disease | 465 (2) | 31 (2) | 496 (2) |
| **BMI measurement** | | | |
| 1 Within 30 days prior to the index date | 15,742 (84) | 1,006 (80) | 16,748 (84) |
| 2 Within 30 days after the index date | 1,353 (7) | 186 (15) | 1,539 (8) |
| 3 Within 180 to 30 days prior to the index date | 511 (3) | 13 (1) | 524 (3) |
| 4 Outside prioritized periods | 636 (3) | 24 (2) | 660 (3) |
| 5 No test | 484 (3) | 21 (2) | 505 (3) |
| **BMI** | | | |
| BMI 0 to <18.5 kg/m2 | 1,432 (8) | 72 (6) | 1,504 (8) |
| BMI 18.5 to <25 kg/m2 | 7,323 (39) | 475 (38) | 7,798 (39) |
| BMI 25 to <30 kg/m2 | 5,183 (28) | 384 (31) | 5,567 (28) |
| BMI 30 to <35 kg/m2 | 2,327 (12) | 185 (15) | 2,512 (13) |
| BMI 35 to <40 kg/m2 | 850 (5) | 63 (5) | 913 (5) |
| BMI >= 40 kg/m2 | 491 (3) | 26 (2) | 517 (3) |
| Not measured | 1,120 (6) | 45 (4) | 1,165 (6) |
| **BMI Non-missing N (%)** | 17,606 (94) | 1,205 (96) | 18,811 (94) |
| **BMI Mean (SD)** | 26 (6) | 26 (6) | 26 (6) |
| **BMI Median (IQR)** | 25 (22-29) | 26 (22-29) | 25 (22-29) |
| **Alcohol intake registration** | | | |
| 1 Within 30 days prior to the index date | 4,699 (25) | 291 (23) | 4,990 (25) |
| 2 Within 30 days after the index date | 1,358 (7) | 101 (8) | 1,459 (7) |
| 3 Within 180 to 30 days prior to the index date | 1,826 (10) | 110 (9) | 1,936 (10) |
| 4 Outside prioritized periods | 5,268 (28) | 359 (29) | 5,627 (28) |
| 5 No test | 5,575 (30) | 389 (31) | 5,964 (30) |
| **Alcohol** | | | |
| Alcohol intake above recommandations | 2,039 (11) | 119 (10) | 2,158 (11) |
| Alcohol intake within recommandations | 5,844 (31) | 383 (31) | 6,227 (31) |
| No reported alcohol intake | 10,843 (58) | 748 (60) | 11,591 (58) |
| **Smoking status registration** | | | |
| 1 Within 30 days prior to the inclusion date | 3,536 (19) | 249 (20) | 3,785 (19) |
| 2 Within 30 days after the inclusion date | 1,262 (7) | 70 (6) | 1,332 (7) |
| 3 Within 180 to 30 days prior to the inclusion date | 2,630 (14) | 148 (12) | 2,778 (14) |
| 4 Outside prioritized periods | 6,323 (34) | 434 (35) | 6,757 (34) |
| 5 No test | 4,975 (27) | 349 (28) | 5,324 (27) |
| **Smoking status** | | | |
| Current smoker | 1,915 (10) | 125 (10) | 2,040 (10) |
| No smoking status reported | 11,298 (60) | 783 (63) | 12,081 (60) |
| Non-smoker | 2,195 (12) | 124 (10) | 2,319 (12) |
| Occational smoker | 91 (0) | 7 (1) | 98 (0) |
| Previous smoker | 3,227 (17) | 211 (17) | 3,438 (17) |
| **Smoking status** | | | |
| No | 13,493 (72) | 907 (73) | 14,400 (72) |
| Yes | 5,233 (28) | 343 (27) | 5,576 (28) |
| **Smoking status any time before hospitalisation** | | | |
| Current smoker | 3,367 (18) | 221 (18) | 3,588 (18) |
| No smoking status reported | 8,022 (43) | 543 (43) | 8,565 (43) |
| Non-smoker | 2,818 (15) | 161 (13) | 2,979 (15) |
| Occational smoker | 152 (1) | 13 (1) | 165 (1) |
| Previous smoker | 4,367 (23) | 312 (25) | 4,679 (23) |
| **Smoking status any time before hospitalisation** | | | |
| No | 10,840 (58) | 704 (56) | 11,544 (58) |
| Yes | 7,886 (42) | 546 (44) | 8,432 (42) |
| **Diastolic blood pressure registration** | | | |
| 1 Within 30 days prior to the inclusion date | 18,558 (99) | 1,220 (98) | 19,778 (99) |
| 2 Within 30 days after the inclusion date | 51 (0) | 28 (2) | 79 (0) |
| 3 Within 180 to 30 days prior to the inclusion date | 41 (0) | 0 (0) | 41 (0) |
| 4 Outside prioritized periods | 47 (0) | 2 (0) | 49 (0) |
| 5 No test | 29 (0) | 0 (0) | 29 (0) |
| Diastolic blood pressure (mmHg) Non-missing N (%) | 18,650 (100) | 1,248 (100) | 19,898 (100) |
| Diastolic blood pressure (mmHg) Mean (SD) | 75 (13) | 72 (14) | 75 (13) |
| Diastolic blood pressure (mmHg) Median (IQR) | 74 (66-83) | 70 (61-80) | 74 (65-83) |
| **Systolic blood pressure registration** | | | |
| 1 Within 30 days prior to the inclusion date | 18,558 (99) | 1,220 (98) | 19,778 (99) |
| 2 Within 30 days after the inclusion date | 51 (0) | 28 (2) | 79 (0) |
| 3 Within 180 to 30 days prior to the inclusion date | 41 (0) | 0 (0) | 41 (0) |
| 4 Outside prioritized periods | 47 (0) | 2 (0) | 49 (0) |
| 5 No test | 29 (0) | 0 (0) | 29 (0) |
| Systolic blood pressure (mmHg) Non-missing N (%) | 18,650 (100) | 1,248 (100) | 19,898 (100) |
| Systolic blood pressure (mmHg) Mean (SD) | 133 (22) | 127 (25) | 133 (23) |
| Systolic blood pressure (mmHg) Median (IQR) | 131 (117-147) | 124 (110-142) | 130 (116-147) |
| **High blood pressure (diastolic > 90 mm Hg or systolic > 140 mm Hg)** | | | |
| No | 11,969 (64) | 889 (71) | 12,858 (64) |
| Yes | 6,757 (36) | 361 (29) | 7,118 (36) |
| **Use of prescription medicine (within 180 days prior to hospitalisation)** | | | |
| Antidiabetics (A10A, A10B) | 3,236 (17) | 252 (20) | 3,488 (17) |
| Blood pressure medication (C02, C07, C08, C09) | 10,281 (55) | 765 (61) | 11,046 (55) |
| Antibiotics (J01) | 9,396 (50) | 607 (49) | 10,003 (50) |
| Inhaled corticosteroid therapy (R03) | 4,524 (24) | 297 (24) | 4,821 (24) |
| Lipid-lowering treatment (C10) | 6,170 (33) | 483 (39) | 6,653 (33) |
| Low dose aspirin (B01AC06 75, 100, 150 mg, N02BA01 100 mg) | 4,887 (26) | 377 (30) | 5,264 (26) |
| **Hospital contacts (within 30 days prior to hospitalisation)** | | | |
| At least one hospital contact | 17,594 (94) | 1,067 (85) | 18,661 (93) |
| At least one inpatient hospital contact | 15,747 (84) | 937 (75) | 16,684 (84) |
| At least one inpatient hospital contact with acute admission | 15,431 (82) | 903 (72) | 16,334 (82) |
| At least one outpatient hospital contact | 6,420 (34) | 416 (33) | 6,836 (34) |
| At least one emergency room visit | 3,613 (19) | 202 (16) | 3,815 (19) |

Data are given as number (percentage) of patients, unless otherwise specified.

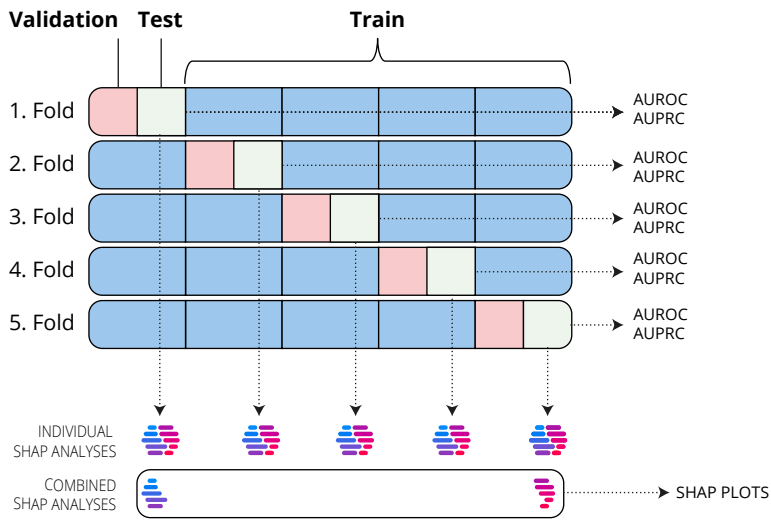

**Supplementary Figure 1 | Cross-validation scheme.**
Data were randomly divided into 5 portions of 20% each. For each fold four portions (80 %) were used to fit the xAI-EWS model parameters during training. The remaining 20% was split into two portions of 10% each for validation and test. The validation data were used to perform an unbiased evaluation of a model fit during training, and the test data were used to provide an unbiased evaluation of the final model. For each fold data were shifted such that a new portion was used for testing. All data for a single patient was assigned to either train, validation or test data. SHAP analysis was conducted for each fold in cross validation, giving a total of five individual SHAP analyses. These five SHAP analyses were combined in one pooled analysis, which was the basis for the explanation results given in Figure 4 and Figure 5.

**Supplementary table 3 - Area under the receiver operating characteristic curve results**

| Fixed time to onset (%) | Sliding window (%) | Sliding window w. D.I (%) | On clinical demand (%) |
|---|---|---|---|
| .400 | .010 | .010 | .013 |
| .390 | .007 | .010 | .012 |
| .348 | .006 | .008 | .017 |
| .385 | .008 | .007 | .013 |
| .390 | .006 | .009 | .016 |

**Supplementary table 4 - Area under the receiver operating characteristic curve derived confidence intervals**

|  | Fixed time to onset (%) | Sliding window (%) | Sliding window w. D.I (%) | On clinical demand (%) |
|---|---|---|---|---|
| Confidence | .0251 | .0019 | .0013 | .0030 |
| Mean | .3825 | .0074 | .0088 | .0143 |
| 95% CI lower limit | .3575 | .0055 | .0075 | .0113 |
| 95% CI lower limit | .4076 | .0093 | .0101 | .0173 |

**Supplementary table 5 - Area under the receiver operating characteristic curve results**

| Fixed time to onset (%) | Sliding window (%) | Sliding window w. D.I (%) | On clinical demand (%) |
|---|---|---|---|
| .8181 | .7651 | .7660 | .7549 |
| .8259 | .7833 | .7577 | .7440 |
| .8101 | .7738 | .7342 | .7863 |
| .8173 | .7817 | .7415 | .7475 |
| .8192 | .7813 | .7238 | .7899 |

**Supplementary table 6 - Area under the receiver operating characteristic curve derived confidence intervals**

|  | Fixed time to onset (%) | Sliding window (%) | Sliding window w. D.I (%) | On clinical demand (%) |
|---|---|---|---|---|
| Confidence | 0070 | .0095 | .0213 | .0272 |
| Mean | .8181 | .7771 | .7446 | .7645 |
| 95% CI lower limit | .8111 | .7676 | .7233 | .7373 |
| 95% CI lower limit | .8251 | .7865 | .7659 | .7917 |

**Code for analysis can be found in the github repository:**
https://github.com/SimonMeyerLauritsen/Framing/blob/main/Analysis%20pipeline.py